\documentclass{article}
\usepackage{spconf,amsmath,epsfig}

\title{Impulse Noise Removal in Speech Using Wavelets}

\name{R.~C.~Nongpiur \thanks{Thanks to QNX Software Systems - Wavemakers for providing some of the resources for conducting the experiments}}
\address{QNX Software Systems - Wavemakers \\
         Vancouver, British Columbia \\
	 Canada.\\rnongpiur@ieee.org}

\begin{document}
\ninept
% make the title area
\maketitle

\begin{abstract}
A new method for removing impulse noise from speech in the \linebreak wavelet transform domain is proposed. The method utilizes the multi-resolution property of the wavelet transform, which provides finer time resolution at the higher frequencies than the short-time Fourier transform (STFT), to effectively identify and remove impulse noise. It uses two features of speech to discriminate speech from impulse noise: one is the slow time-varying nature of speech and the other is the Lipschitz regularity of the speech components. On the basis of these features, an algorithm has been developed to identify and suppress wavelet coefficients that correspond to impulse noise. Experiment results show that the new method is able to significantly reduce impulse noise without degrading the quality of the speech signal or introducing any audible artifacts.
\end{abstract}


\section{Introduction}
The presence of impulse-like noise in speech can significantly reduce the intelligibility of speech and degrade automatic speech \linebreak recognition (ASR) performance. Impulse noise is characterized by a short burst of acoustic energy of either a single impulse or a series of impulses, with a wide spectral bandwidth. Typical acoustic impulse noises include sounds of machine gun firing, of rain drops hitting a hard surface like the windshield of a moving car, of typing on a keyboard, of indicator clicks in cars, of clicks in old analog recordings, of popping popcorn and so on. One difficulty with removing impulse noise from speech is the wide temporal and spectral variation between different parts of speech, such as the periodic and low-frequency nature of vowel signals and the random and high-frequency nature of consonants. An effective algorithm should, therefore, consistently remove the impulse noise whether it falls in vowels, consonants, or silent portions of speech. For audio signals, several time domain algorithms have been developed to detect and remove impulse noise ~\cite{vaseghi1, esquef, chandra}. However, since these algorithms work in the time domain they do not effectively utilize the frequency information of the signals. Further, classical block processing methods such as the STFT algorithm or the linear prediction (LP) algorithm have also been used ~\cite{liu, vaseghi2} to detect or remove impulse-like sounds; however, two problems may result if classic block processing techniques are used: the first is determining the exact position of the impulse within the analyzed data-frame -- these methods give no straightforward information about the position of the impulse within the analyzed frame. It is possible, however, to reduce the frame size to achieve better resolution in time; but doing this leads to the second problem where we lose the frequency resolution to analyze the signal. The wavelet transform overcomes both the difficulties due to its multi-resolution property ~\cite{mallatbook}. In multi-resolution analysis, the window length or wavelet scale for analyzing the frequency components increases as the frequency decreases. This property enables the wavelet transform to have better time resolution for higher frequency components and better frequency resolution for lower ones. Consequently, by using the wavelet transform we have the benefit of both time and frequency resolutions to detect and remove impulse noise. In this paper, we utilize the slow time-varying nature of speech relative to the duration of an impulse to detect and suppress the impulses at finer scales. On the basis of the impulse detected in the finer scales, the corresponding wavelet coefficients of the impulse at the coarser scales are attenuated. Our experiment results show that the impulse noise is significantly suppressed without degrading the speech signal or introducing any audible artifacts.
 The paper is organized as follows. Section 2 discusses the difference in smoothness, or regularity, between speech and impulse noise. In Section 3, the removal of impulse noise from speech in the wavelet transform domain is described. In Section 4, simulation results are presented to show the effectiveness of the proposed approach. Conclusions are drawn in Section 5.

\section{Regularity of speech and impulse noise}
The regularity of a signal is defined as the number of continuous derivatives that the signal possesses. It can be estimated from the Holder exponent, also know as the Lipschitz exponent ~\cite{hwang}, which is defined as follows: If we assume that a signal $f(t)$ can be approximated locally at $v$ by a polynomial of the form
\begin{equation}
\begin{split}
f(t) & = c_0+c_1(t-v)+\cdots+c_n(t-v)^n+C|t-v|^\alpha \\
     & = p_n(t-v) + C|t-v|^\alpha
\end{split}
\label{eq:1}
\end{equation}
where $p_n$ is a polynomial of order $n$ and $C$ is a coefficient, then the term associated with Lipschitz exponent, $\alpha$, is the part of the signal that does not fit into the $n+1$ approximation ~\cite{mallatbook}. The local regularity of a function at $v$ can be characterize by $\alpha$ such that
\begin{equation}
|f(t) - p_n(t-v)| \leq C |t-v|^\alpha
\label{eq:2}
\end{equation}
where a higher value of $\alpha$ indicates a better regularity or smoother function. In ~\cite{hwang}, it is shown that a bounded function $f(x)$ is uniformly Lipschitz $\alpha$ over an interval ${\bf{R}}$ if it satisfies
\begin{equation}
\int_{-\infty}^{+\infty} |F(w)|(1+|w|^\alpha)dw < +\infty
\label{eq:3}
\end{equation}
where $F(w)$ is the Fourier transform of $f(w)$. The condition gives the global regularity for the whole real line and implies that $F(w)$ has a decay faster than $1/w^\alpha$. 

A speech signal can be considered to be broadly made up of vowels, consonants and silence portions. The vowel portion is generated by vibrations in the vocal chords which are then lowpass filtered by the vocal tract. As such, vowels are usually periodic with an upper cutoff frequency that does not exceed 5 kHz. The consonants, on the other hand, are generated by constriction in the mouth; they are usually random with a spectrum that can extend up to 20 kHz. If we use condition (\ref{eq:3}), it is clear that a vowel due to its lower bandwidth will have larger positive Lipschitz value than a consonant which may even be negative. The silence portion of speech is essentially background noise that is random with a small or negative Lipschitz value. Consequently, by knowing the random or periodic nature of different parts of speech and their corresponding regularity, it is possible to make a better decision when removing impulse noise from speech. Another important characteristics of speech, is the slow time varying nature of the temporal and spectral envelop in comparison to an impulse; this slow-time varying nature is because speech is generated by the movements of muscles in the mouth and vocal tract, which is a relatively slow process. 
 
A higher value of $\alpha$ indicates a better regularity or a smoother function. To detect an impulse, we need a transform that ignores the polynomial part in (\ref{eq:1}). A wavelet transform $\psi(x)$ with $n$ vanishing moments is able to ignore a polynomial upto order $n$. Such a wavelet satisfies the condition
\begin{equation}
\int_{-\infty}^{+\infty} t^n \psi(t) dt = 0
\label{eq:4}
\end{equation}
and using it to transform (\ref{eq:2}) we get the inequality
\begin{equation}
|Wf(s, t)| \leq C s^\alpha
\label{eq:5}
\end{equation}
where $Wf(s,t)$ is the wavelet transform of $f(t)$ at scale $s$ ~\cite{hwang}. To estimate the local regularity of a signal we set the inequality in (\ref{eq:5}) to equality and take the logarithm on both sides:
\begin{equation}
\log{|Wf(t, s)|} = \log(C) + \alpha \log{s}
\label{eq:6}
\end{equation}
 The Lipschitz value is simply the slope of the decay of the wavelet coefficients across scales, given by 
\begin{equation}
\alpha = \frac{\Delta \log|Wf(t, s)|}{\Delta \log{s}}
\label{eq:7}
\end{equation}

An impulse is characterized by a sudden change in the signal or a sudden shift in the signal mean value. Large magnitude coefficients, termed modulus maxima, will be present at time points where the impulses have occurred. Impulses are distinguishable from noise by the presence of modulus maxima at all of the scale levels; noise, on the other hand, produces modulus maxima only at finer scales. In ~\cite{hwang}, Mallat and Hwang used a method for detecting singularities by analyzing the evolution of the wavelet modulus maxima across scales, for a continuous wavelet transform; the decay of the maxima line was used to determine the regularity at a given point. However, in practical application a dyadic discrete wavelet transform is preferred over the continuous wavelet transform due to its lower computational effort. Therefore, if a discrete wavelet transform is used the number of wavelet coefficients will reduce by half for the next increase in scale. In such a case, rather than extracting the maxima line it is computationally more efficient to simply estimate the decay of the wavelet coefficients for each time point in the smallest scale. Points where large changes occur in the signal, such as an impulse, would have large coefficients at all the different scales, thus having little decay across scales. Noise, on the other hand, would have large coefficients only at finer scales and would, therefore, show faster decay towards coarser scales. Likewise, the consonants in speech being similar to high-pass filtered white noise would also show more decay towards coarser scales; however, vowels, with frequency spectrum that does not exceed 5 kHz, are characterized by small coefficients in the smallest scale that increases as the scale increases.

\section{Removal of impulse noise from speech}
Since our objective is the removal of impulse noise from speech, it is not critical that impulses of lower magnitudes that are perceptually inaudible be removed. It is important, however, that impulses of larger magnitudes be suppressed below the just-noticeable level difference (JNLD) to make them inaudible. The JNLD is not a fixed value and varies with the nature of the signal and sound pressure level (SPL). For example, for white noise the JNLD is around 0.7 dB for SPL between 40 and 100 dB, while for a 1 kHz tone the JNLD decreases from 1 to 0.2 dB as the SPL increases from 40 to 100 dB ~\cite{zwicker}. Consequently, an impulse would be perceptually inaudible for most SPL levels if it is suppressed below 0.2 dB above the speech signal level. 

As discussed in Section 2, the temporal and spectral envelop of speech is slow time-varying in comparison to an impulse. This property is used to detect and suppress the wavelet coefficients that correspond to an impulse. Therefore, what is needed is a dynamic threshold for each wavelet level that varies in proportion to the smooth envelop of the absolute wavelet coefficients values, but, at the same time, not affected by impulse noise. That is, for scale $s$ and sample $n$, such a dynamic threshold, $\tau(s, n)$, can be defined as
\begin{equation}
\tau(s, n) = k_s \cdot \mbox{Env}\bigl[|Wf(n, s)|\bigr]
\label{eq:8}
\end{equation}
where operator $\mbox{Env}[\cdot]$ is the envelop of the signal that is unaffected by impulse noise and $k_s$ is a factor that is determined empirically for each level on the basis of the JNLD and the nature of the impulse noise. A median filter is known to possess the property where step-function type signals are preserved while at the same time robust to impulse noise ~\cite{yang}. As such, the operator $\mbox{Env}[\cdot]$ in (\ref{eq:8}) can be replaced by a median filter of length $N=2K+1$ so that $\tau(s, n)$ becomes
\begin{equation}
\begin{split}
\tau(s, n) & = k_s \cdot \mbox{MED}\Bigl[ |Wf(n-K,s)|, \cdots,\Bigr. \\
& \quad \Bigl. |Wf(n, s)|, \cdots, |Wf(n+K,s)| \Bigr]
\end{split}
\label{eq:9}
\end{equation}
The length of the median filter needs be adjusted so that it is sufficiently long in comparison to an impulse but short in comparison to a vowel or consonant. A wavelet coefficient would be considered to be that of an impulse if it greater than $\tau(s, n)$; to suppress the impulse the coefficient is attenuated to a new coefficient, $\widehat{Wf}(n,s)$, given by 
\begin{equation}
\widehat{Wf}(n, s) = \begin{cases} 
Wf(n, s) & \text{if $|Wf(n, s)| < \tau(s, n)$}, \\
\tau(s, n)\frac{Wf(n, s)}{|Wf(n, s)|} & \text{otherwise}.
\end{cases}
\label{eq:9a}
\end{equation}
For impulses that occur in a consonant or in the non-voice portion of speech, suppression of the wavelet coefficients at coarser scales is as important as in the finer scales; this is because an impulse has a Lipschitz exponent that is usually greater than a consonant or background noise and, as such, its coefficients will not decay as fast at coarser scales. And if these coefficients at coarser scales are not suppressed adequately, they will be audible as low frequency thuds. However, for an impulse that occurs in the middle of a vowel, suppression of the corresponding coefficients at coarser scales is not as critical as in the finer scales. This is because the Lipschitz exponent of a vowel is much greater than that of an impulse, and at coarser scales the contribution to the coefficients comes mainly from the vowel. This also implies that the vowel will usually mask out the low-frequency portion of an impulse.

At larger scales, the use of (\ref{eq:8}) and (\ref{eq:9}) to detect the coefficients that correspond to an impulse become less effective if a discrete wavelet transform is used since the number of sample points decreases by half for the next increase in wavelet scale, thereby reducing the time resolution by half; furthermore, at larger scales an impulse will have much lesser contributions on a wavelet coefficient since other portions of the signal within the wavelet support length will also contribute to the coefficient. Therefore, for wavelet coefficients at larger scales it becomes more effective if the attenuation is done on the basis of the impulses detected in the smaller scales. If $\lambda_I(s)$ is the average decay of an impulse and $\lambda_I(s_0) = 1$, where $s_0$ is the finest scale, then the attenuated wavelet coefficient, $\widehat{Wf}(s_c, n)$, for the coarser scale, $s_c$, is given by
\begin{equation}
\widehat{Wf}(s_c, n) = \begin{cases}
0 & \text{if $K < 0$}, \\
K & \text{otherwise}.
\end{cases}
\label{eq:10}
\end{equation}
where
\begin{equation}
K = Wf(s_c, n) - \frac{k_c\lambda_I(s_c)}{\lambda_I(s_f)} \bigl[|Wf(s_f, n)| - \tau(s_f, n)\bigr],
\label{eq:11}
\end{equation}
$|Wf(s_f, n)|$ is the absolute wavelet coefficient of the detected impulse in the finer scale $s_f$, and $k_c$ is a constant that is empirically determined depending on the type of impulse noise and the JNLD at that scale.

\section{Experimental Results}
In this section we show the effectiveness of our proposed method by cleaning speech that is corrupted by a certain type of impulse noise, namely, the sound of raindrops hitting the windshield and roof of a moving car. In the automotive environment, the removal of the sound of falling raindrops from speech is important for achieving high-clarity speech for automatic speech recognition and hands-free systems in a car during rain.
  
For our experiments, we use rain sound that has been recorded at 16 kHz sampling rate by an AKG-Q400 microphone placed on the mirror of a car that is moving at 60 km/hr. The clean speech signal is taken from the TIMIT database; it is first convolved with an impulse-response that simulates the channel between the driver's mouth and the car microphone before being mixed with the recorded rain-sound. The mixing ratio between the clean speech signal and impulse noise is adjusted to a realistic car-environment level. To process the signal, the speech signal is divided into blocks each of length 512 with 50 percent overlap. The Daubechies wavelet transform of order 6 is applied on each of the blocks for analysis; after that, we apply our proposed method to suppress wavelet coefficients that correspond to rain impulses. Finally, we apply the inverse wavelet transform on the modified wavelet coefficients to transform it back to the time domain. To remove edge effects when combining the time-domain blocks, a Hanning window is applied on each of the blocks before performing the overlap-and-add. Once the impulses have been suppressed standard noise removal techniques ~\cite{boll, mallah} may be applied to enhance the speech signal further.

In Fig.~1(a), we illustrate the typical spectrogram of a vowel, a consonant, and a rain impulse, and in Fig.~1(b) we plot their corresponding decay rates after normalizing the value at the finest scale to unity. As can be seen in Fig.~1(b), the consonant has the fastest decay followed by the rain impulse and then the vowel. Though different vowels and consonants will have variations in their decay rates, their decay trend will be similar to the curves in Fig.~1(b). From Fig.~1(b), it is apparent that an impulse that occurs in a consonant will need to be adequately suppressed even at coarser scales, while an impulse that occurs in a vowel will require suppression only at the finer scales. The average decay of the rain impulse, $\lambda_I(s)$, which is required in (\ref{eq:10}) for suppressing the coefficients of impulses at coarser scales, is determined by taking the average of the normalized decay of a sufficient number of typical rain impulses. 
\begin{figure}[htb]
\begin{minipage}[b]{1.0\linewidth}
  \centering
  \centerline{\epsfig{figure=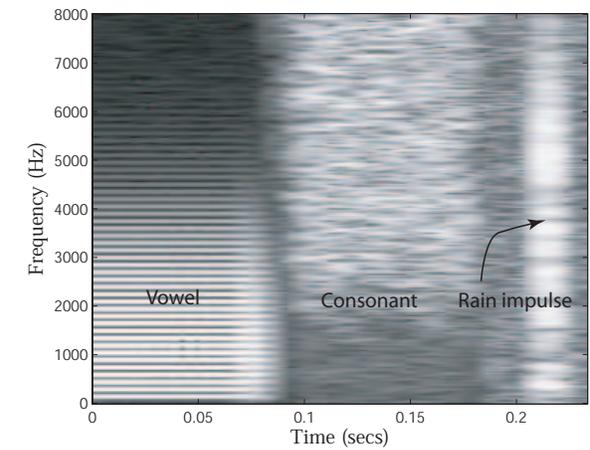,width=8.5cm}}
%  \vspace{2.0cm}
  \centerline{(a)}\medskip
\end{minipage}
\begin{minipage}[b]{1.0\linewidth}
  \centering
  \centerline{\epsfig{figure=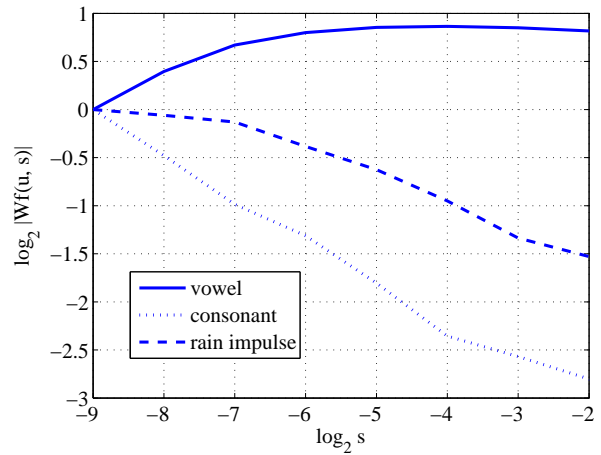,width=8.5cm}}
%  \vspace{1.5cm}
  \centerline{(b)}\medskip
\end{minipage}
\caption{(a) A typical spectrogram of a vowel, a consonant, and a rain impulse; (b) their corresponding normalized wavelet magnitude variation across scales.}
\label{fig:res}
\end{figure}
To suppress the wavelet coefficients that correspond to the rain impulses, we use equation (\ref{eq:9a}) for the 3 finest scales and (\ref{eq:10}) for the 4th, 5th, and 6th finest. Spectrograms plots of speech before and after removal of the rain impulses are illustrated in Figs.~2(a) and (b), respectively. Audio examples may be accessed from the following website: www.ece.uvic.ca/\textasciitilde rnongpiu/audio.html.

Though a performance analysis with different types of wavelets is beyond the scope of this paper, it must be mentioned that the type of wavelet used and the number of vanishing moments will also affect the ability to discriminate impulse noise from speech. For example, if a wavelet with a large vanishing moment is used, we may get better performance when separating an impulse from a vowel but lower performance when separating it from a consonant. Furthermore, by designing wavelets that are specifically tuned for discriminating speech from the impulse noise in consideration, it is possible to get even more improvement. 

\begin{figure}[htb]
\begin{minipage}[b]{1.0\linewidth}
  \centering
  \centerline{\epsfig{figure=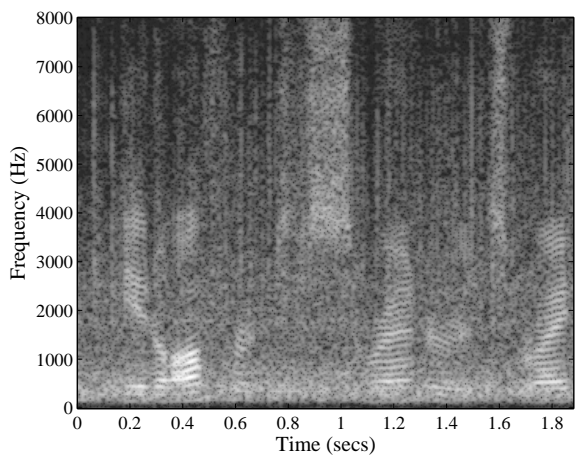,width=8.5cm}}
%  \vspace{2.0cm}
  \centerline{(a)}\medskip
\end{minipage}
\begin{minipage}[b]{1.0\linewidth}
  \centering
  \centerline{\epsfig{figure=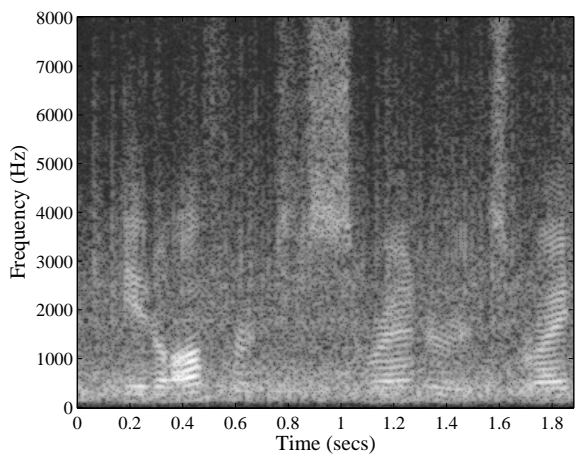,width=8.5cm}}
%  \vspace{1.5cm}
  \centerline{(b)}\medskip
\end{minipage}
\caption{(a) Spectogram of speech corrupted with sounds of raindrops hitting the windshield and roof of a car, which is moving at 60 km/hr; (b) Spectogram of speech after suppression of the rain impulses by the proposed algorithm.}
\label{fig:res}
\end{figure}

\section{Conclusions}
A new method for removing impulse noise from speech in the wave-\linebreak let transform domain has been described. The method utilized the multi-resolution property of the wavelet transform, which provides finer time resolution at high frequencies, to effectively identify and remove the impulse noise. To discriminate the impulse from speech it uses the slow time-varying nature of speech relative to an impulse, and the difference in regularity between an impulse and various parts of speech. On the basis of these differences, an algorithm was developed to identify and suppress wavelet coefficients that correspond to impulse noise. Experiment results have shown that the new method is able to significantly reduce impulse noise without degrading the quality of speech signal or introducing any artifacts.

\end{document}